\documentclass[sigconf]{acmart}
\usepackage{booktabs}

\usepackage{hyperref}
\usepackage{multirow}
\usepackage{pbox}
\usepackage{todonotes}

\usepackage{url}            
\usepackage{booktabs}       

\usepackage[utf8]{inputenc} 
\usepackage[T1]{fontenc}    
\usepackage{hyperref}       
\usepackage{amsfonts}       
\usepackage{nicefrac}       
\usepackage{microtype}      
\usepackage{graphicx} 
\usepackage{caption}
\usepackage{subcaption}
\usepackage{amsmath}
\usepackage[linesnumbered,ruled]{algorithm2e}
\usepackage{color,soul}

\setcopyright{rightsretained}

\begin{document}

\copyrightyear{2017} 
\acmYear{2017} 
\setcopyright{acmcopyright}
\acmConference{KDD'17}{}{August 13-17, 2017, Halifax, NS, Canada}
\acmPrice{15.00}\acmDOI{10.1145/3097983.3098177}
\acmISBN{978-1-4503-4887-4/17/08}

\fancyhead{}
\settopmatter{printacmref=false}

\title{ReasoNet: Learning to Stop Reading in Machine Comprehension}

%

\author{Yelong Shen, Po-Sen Huang, Jianfeng Gao, Weizhu Chen}
\affiliation{%
  \institution{Microsoft Research}
  \streetaddress{One Microsoft Way}
  \city{Redmond} 
  \state{WA} 
  \postcode{98053}
}
\email{yeshen, pshuang, jfgao, wzchen@microsoft.com}

\begin{abstract}

Teaching a computer to read and answer general questions pertaining to a document is a challenging yet unsolved problem. In this paper, we describe a novel neural network architecture called the Reasoning Network ({ReasoNet}) for machine comprehension tasks. ReasoNets make use of multiple turns to effectively exploit and then reason over the relation among queries, documents, and answers. Different from previous approaches using a fixed number of turns during inference, ReasoNets introduce a termination state to relax this constraint on the reasoning depth. With the use of reinforcement learning, ReasoNets can dynamically determine whether to continue the comprehension process after digesting intermediate results, or to terminate reading when it concludes that existing information is adequate to produce an answer. ReasoNets achieve superior performance in machine comprehension datasets, including unstructured CNN and Daily Mail datasets, the Stanford SQuAD dataset, and a structured Graph Reachability dataset. 


\end{abstract}



\maketitle

\section{Introduction}
Teaching machines to read, process, and comprehend natural language documents is a coveted goal for artificial intelligence \citep{bottou2014machine, richardson2013mctest, HermannNIPS2015}. Genuine reading comprehension is extremely challenging, since effective comprehension involves thorough understanding of documents and sophisticated inference. 
Toward solving this machine reading comprehension problem, in recent years, several works have collected various datasets, in the form of question, passage, and answer, to test machine on answering a question based on the provided passage \citep{richardson2013mctest, HermannNIPS2015, HillICLR2016, Squad2016}. Some large-scale cloze-style datasets \citep{HermannNIPS2015, HillICLR2016} have gained significant attention along with powerful deep learning models. 

Recent approaches on cloze-style datasets can be separated into two categories: single-turn and multi-turn reasoning. Single turn reasoning models utilize attention mechanisms \citep{bahdanau2015attention} to emphasize specific parts of the document which are relevant to the query. These attention models subsequently calculate the relevance between a query and the corresponding weighted representations of document subunits (e.g. sentences or words) to score target candidates \citep{HillICLR2016, HermannNIPS2015, KadlecAttentionSum2016}. However, considering the sophistication of the problem, after a single-turn comprehension, readers often revisit some specific passage or the question to grasp a better understanding of the problem. With this motivation, recent advances in reading comprehension have made use of multiple turns to infer the relation between query, document and answer \citep{HillICLR2016, Dhingra2016Gated, Trischler2016EpiReader, Sordoni2016IterativeAtt}. By repeatedly processing the document and the question after digesting intermediate information, multi-turn reasoning can generally produce a better answer and these existing works have demonstrated its superior performance consistently.

Existing multi-turn models have a pre-defined number of hops or iterations in their inference without regard to the complexity of each individual query or document. However, when human read a document with a question in mind, we often decide whether we want to stop reading if we believe the observed information is adequate already to answer the question, or continue reading after digesting intermediate information until we can answer the question with confidence. This behavior generally varies from document to document or question to question because it is related to the sophistication of the document or the difficulty of the question. Meanwhile, the analysis in \cite{ChenACL2016} also illustrates the huge variations in the difficulty level with respect to questions in the CNN/Daily Mail datasets \citep{HermannNIPS2015}. For a significant part of the datasets, this analysis shows that the problem cannot be solved without appropriate reasoning on both its query and document.


With this motivation, we propose a novel neural network architecture called Reasoning Network ({ReasoNet}). which tries to mimic the inference process of human readers. With a question in mind, ReasoNets read a document repeatedly, each time focusing on different parts of the document until a satisfying answer is found or formed. This reminds us of a Chinese proverb: \emph{``The meaning of a book will become clear if you read it hundreds of times.''}.
Moreover, unlike previous approaches using fixed number of hops or iterations, ReasoNets introduce a termination state in the inference. This state can decide whether to continue the inference to the next turn after digesting intermediate information, or to terminate the whole inference when it concludes that existing information is sufficient to yield an answer. The number of turns in the inference is dynamically modeled by both the document and the query, and can be learned automatically according to the difficulty of the problem. 

One of the significant challenges ReasoNets face is how to design an efficient training method, since the termination state is discrete and not connected to the final output.  This prohibits canonical back-propagation method being directly applied to train ReasoNets.  Motivated by \cite{williams1992reinforce, mnih2014recurrent}, we tackle this challenge by proposing a reinforcement learning approach, which utilizes an instance-dependent reward baseline, to successfully train ReasoNets. 
Finally, by accounting for a dynamic termination state during inference and applying proposed deep reinforcement learning optimization method, ReasoNets achieve the state-of-the-art results in machine comprehension datasets, including unstructured CNN and Daily Mail datasets, and the proposed structured Graph Reachability dataset, when the paper is first publicly available on arXiv.\footnote{https://arxiv.org/abs/1609.05284} 
At the time of the paper submission, we apply ReasoNet to the competitive Stanford Question Answering Dataset(SQuAD), ReasoNets outperform all existing published approaches and rank at second place on the test set leaderboard.\footnote{http://www.stanford-qa.com} 

This paper is organized as follows. In Section \ref{sec:previous_work}, we review and compare recent work on machine reading comprehension tasks. In Section \ref{sec:reasoning_net}, we introduce our proposed ReasoNet model architecture and training objectives. Section \ref{sec:exp} presents the experimental setting and results on unstructured and structured machine reading comprehension tasks .

\section{Related Work}
\label{sec:previous_work}
Recently, with large-scale datasets available and the impressive advance of various statistical models, machine reading comprehension tasks have attracted much attention. Here we mainly focus on the related work in cloze-style datasets \citep{HermannNIPS2015, HillICLR2016}. Based on how they perform the inference, we can classify their models into two categories: single-turn and multi-turn reasoning.

{\bf Single-turn reasoning}:
Single turn reasoning models utilize an attention mechanism to emphasize some sections of a document which are relevant to a query. This can be thought of as treating some parts unimportant while focusing on other important ones to find the most probable answer. 
Hermann et al. \cite{HermannNIPS2015} propose the attentive reader and the impatient reader models using neural networks with an attention over passages to predict candidates.
Hill et al. \cite{HillICLR2016} use attention over window-based memory, which encodes a window of words around entity candidates, by leveraging an end-to-end memory network \citep{sukhbaatar2015end2endmn}.
Meanwhile, given the same entity candidate can appear multiple times in a passage, Kadlec et al. \cite{KadlecAttentionSum2016} propose the attention-sum reader to sum up all the attention scores for the same entity.  This score captures the relevance between a query and a candidate.  
Chen et al. \cite{ChenACL2016} propose using a bilinear term similarity function to calculate attention scores with pretrained word embeddings.
Trischler et al. \cite{Trischler2016EpiReader} propose the EpiReader which uses two neural network structures: one extracts candidates using the attention-sum reader; the other reranks candidates based on a bilinear term similarity score calculated from query and passage representations.

{\bf Multi-turn reasoning}:
For complex passages and complex queries, human readers often revisit the given document in order to perform deeper inference after reading a document. Several recent studies try to simulate this revisit by combining the information in the query with the new information digested from previous iterations \citep{HillICLR2016, Dhingra2016Gated, Sordoni2016IterativeAtt, Weissenborn2016, KumarMetamindICML2016}.
Hill et al. \cite{HillICLR2016} use multiple hops memory network to augment the query with new information from the previous hop. Gated Attention reader \citep{Dhingra2016Gated} is an extension of the attention-sum reader with multiple iterations by pushing the query encoding into an attention-based gate in each iteration.  
Iterative Alternative (IA) reader \citep{Sordoni2016IterativeAtt} produces a new query glimpse and document glimpse in each iteration and utilizes them alternatively in the next iteration. Cui et al. \cite{Cui2016AoA} further propose to extend the query-specific attention to both query-to-document attention and document-to-query attention, which is built from the intermediate results in the query-specific attention.
By reading documents and enriching the query in an iterative fashion, multi-turn reasoning has demonstrated their superior performance consistently.

Our proposed approach explores the idea of using both attention-sum to aggregate candidate attention scores and multiple turns to attain a better reasoning capability. Unlike previous approaches using a fixed number of hops or iterations, motivated by \cite{nogueira2016nips, mnih2014recurrent}, we propose a termination module in the inference. The termination module can decide whether to continue to infer the next turn after digesting intermediate information, or to terminate the whole inference process when it concludes existing information is sufficient to yield an answer. The number of turns in the inference is dynamically modeled by both a document and a query, and is generally related to the complexity of the document and the query.

\section{Reasoning Networks}
\label{sec:reasoning_net}
\begin{figure}[t]
	\hspace{-5mm}
  \includegraphics[width=0.5\textwidth]{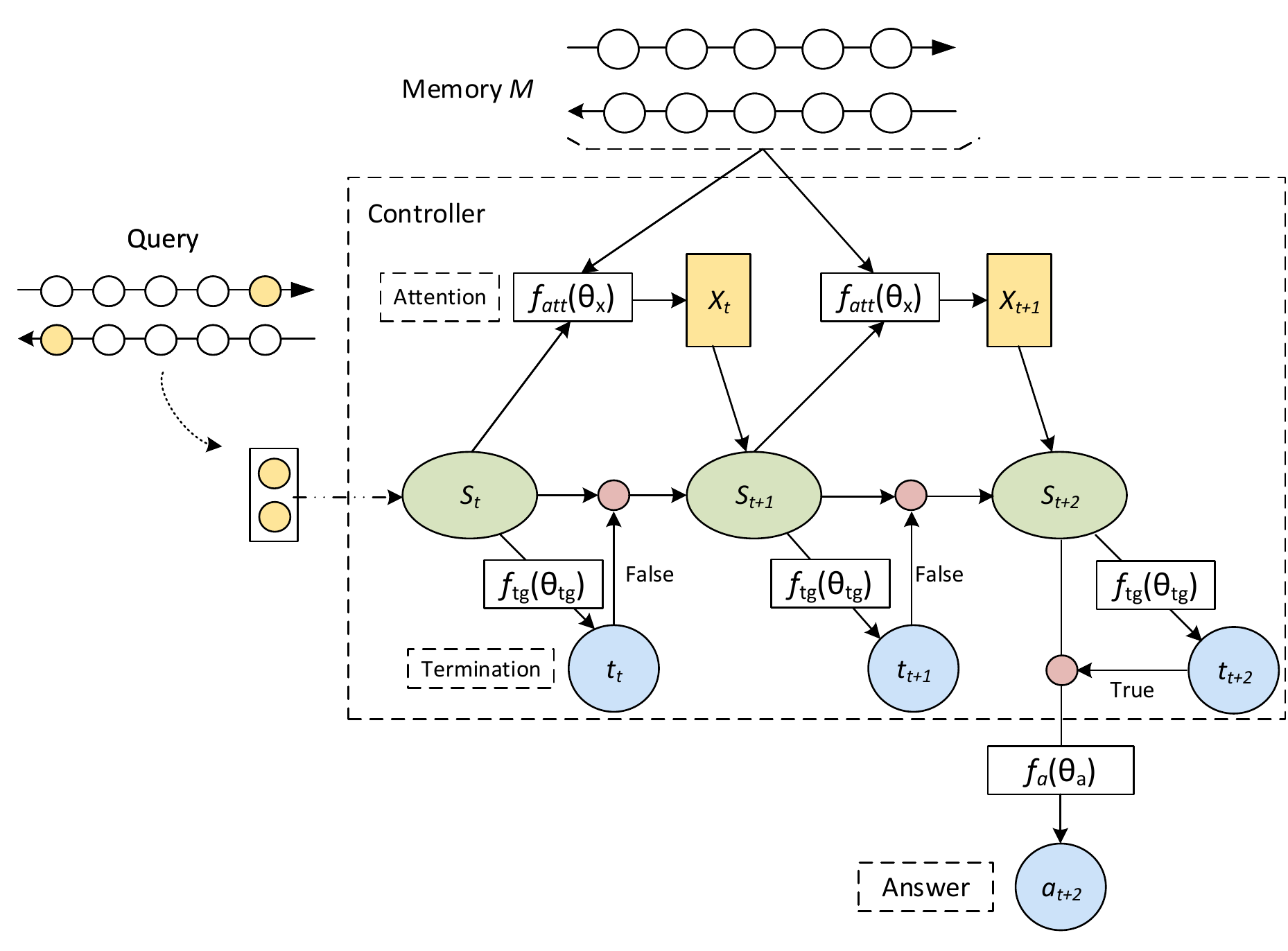} 
  \caption{{A ReasoNet architecture with an episode of \{$t_1=0$, $\ldots$, $t_{t+1}=0$, $t_{t+2}=1$\}}}
  \label{fig:reasonet}
\end{figure}

ReasoNets are devised to mimic the inference process of human readers. ReasoNets read a document repeatedly with attention on different parts each time until a satisfying answer is found. As shown in Figure \ref{fig:reasonet}, a ReasoNet is composed of the following components:


\textbf{Memory}: The external memory is denoted as $M$. It is a list of word vectors, $M = \{m_i \}_{i=1..D}$, where $m_i$ is a fixed dimensional vector. For example, in the Graph Reachability, $m_i$ is the vector representation of each word in the graph description encoded by a bidirectional-RNN. Please refer to Section \ref{sec:exp} for the detailed setup in each experiment.

\textbf{Attention}: The attention vector $x_t$ is generated based on the current internal state $s_t$ and the external memory $M$: $x_t = f_{att}(s_t, M; \theta_x)$. Please refer to Section \ref{sec:exp} for the detailed setup in each experiment.

\textbf{Internal State}: The internal state is denoted as $s$ which is a vector representation of the question state. Typically, the initial state $s_1$ is the last-word vector representation of query by an RNN. The $t$-th time step of the internal state is represented by $s_t$. The sequence of internal states are modeled by an RNN: 
$s_{t+1} = \text{RNN}(s_t, x_t; \theta_s)$, 
where $x_t$ is the attention vector mentioned above. 


\textbf{Termination Gate}: The termination gate generates a random variable according to the current internal state; $t_t \sim p( \cdot |f_{tg}(s_t; \theta_{tg})))$. $t_t$ is a binary random variable. If $t_t$ is true, the ReasoNet stops, and the answer module executes at time step $t$; otherwise the ReasoNet generates an attention vector $x_{t+1}$, and feeds the vector into the state network to update the next internal state $s_{t+1}$.

\textbf{Answer}: The action of answer module is triggered when the termination gate variable is true: $a_t \sim p( \cdot | f_a(s_t ; \theta_a) )$.

\begin{algorithm}[t]
	\SetKwInOut{Input}{Input}
	\SetKwInOut{Output}{Output}
	\Input{Memory $M$; Initial state $s_1$; Step $t = 1$; Maximum Step $T_{\text{max}}$}
	\Output{Termination Step $T$, Answer $a_T$}
	Sample $t_t$ from the distribution $p(\cdot|f_{tg}(s_t; \theta_{tg}))$\;
	if $t_t$ is false, go to Step 3; otherwise Step 6\;
	Generate attention vector $x_t = f_{att}(s_t, M; \theta_{x})$\;
	Update internal state $s_{t+1} = \text{RNN}(s_t, x_t; \theta_s)$\;
	Set $t = t + 1$; if $t < T_{\text{max}}$ go to Step 1; otherwise Step 6\;
	Generate answer $a_t \sim p(\cdot|f_a(s_t; \theta_a))$\;
	Return $T = t$ and $a_T = a_t$\;
	\caption{Stochastic Inference in a ReasoNet}
	\label{alg:inference_process}
\end{algorithm}


In Algorithm \ref{alg:inference_process}, we describe the stochastic inference process of a ReasoNet. The process can be considered as solving a Partially Observable Markov Decision Process (POMDP) \citep{Kaelbling98planningand} in the reinforcement learning (RL) literature. The state sequence $s_{1:T}$ is hidden and dynamic, controlled by an RNN sequence model. The ReasoNet performs an answer action $a_T$ at the $T$-th step, which implies that the termination gate variables $t_{1:T} = (t_1 = 0, t_2 = 0, ..., t_{T-1} = 0, t_{T} = 1)$. The ReasoNet learns a stochastic policy $\pi( (t_t, a_t) | s_t; \theta)$ with parameters $\theta$ to get a distribution of termination actions, to continue reading or to stop, and of answer actions if the model decides to stop at the current step. The termination step $T$ varies from instance to instance.

The learnable parameters $\theta$ of the ReasoNet are the embedding matrices $\theta_W$, attention network $\theta_x$, the state RNN network $\theta_s$, the answer action network $\theta_a$, and the termination gate network $\theta_{tg}$. The parameters $\theta = \{\theta_W, \theta_x, \theta_s, \theta_a, \theta_{tg}\}$ are trained by maximizing the total expect reward. The expected reward for an instance is defined as:
\[ J(\theta) = \mathbb{E}_{\pi(t_{1:T}, a_T; \theta)} \left[ \sum_{t=1}^{T} r_t \right] \]

The reward can only be received at the final termination step when an answer action $a_T$ is performed. We define $r_{T} = 1$ if $t_T = 1$ and the answer is correct, and $r_T = 0$ otherwise. The rewards on intermediate steps are zeros, $\{r_t =0 \}_{t=1 \dots T-1}$.
$J$ can be maximized by directly applying gradient based optimization methods. The gradient of $J$ is given by:
\begin{equation*}
\begin{aligned}
\nabla_{\theta} J(\theta) = \mathbb{E}_{\pi(t_{1:T}, a_T; \theta)} \left[ \nabla_{\theta} \text{log} \pi(t_{1:T}, a_T; \theta)  r_T \right]
\end{aligned}
\end{equation*}

Motivated by the REINFORCE algorithm \citep{williams1992reinforce}, we compute $\nabla_{\theta} J(\theta)$:

\[ \mathbb{E}_{\pi(t_{1:T}, a_T; \theta)} \left[ \nabla_{\theta} \text{log} \pi(t_{1:T}, a_T; \theta)  r_T \right]  = \]
\[ \sum_{(t_{1:T}, a_T) \in \mathbb{A}^\dagger} \pi(t_{1:T}, a_T; \theta)  \left[ \nabla_{\theta} \text{log} \pi(t_{1:T}, a_{T}; \theta) ( r_{T} - b_{T}) \right]   \]
where $\mathbb{A}^\dagger$ is all the possible episodes, $T, t_{1:T}, a_{T} $ and $r_{T}$ are the termination step, termination action, answer action, and reward, respectively, for the ($t_{1:T}$, $a_T$) episode. $b_{T}$ is called the reward baseline in the RL literature to lower the variance \citep{SuttonPhDThesis}. It is common to select $b_{T} = \mathbb{E}_{\pi}\left[r_{T} \right]$ \citep{sutton2001policy}, and can be updated via an online moving average approach : $b_{T} = \lambda b_{T} + (1- \lambda) r_{T}$.
However, we empirically find that the above approach leads to slow convergence in training ReasoNets. Intuitively, the average baselines $\{ b_{T} ; T=1..T_\text{max} \}$ are global variables independent of instances. It is hard for these baselines to capture the dynamic termination behavior of ReasoNets. Since ReasoNets may stop at different time steps for different instances, the adoption of a global variable without considering the dynamic variance in each instance is inappropriate. To resolve this weakness in traditional methods and account for the dynamic characteristic of ReasoNets, we propose an instance-dependent baseline method
to calculate $\nabla_{\theta} J(\theta)$, as illustrated in Section \ref{sec:train_detail}.
Empirical results show that the proposed reward schema achieves better results compared to baseline approaches.


\subsection{Training Details}
\label{sec:train_detail}
In the machine reading comprehension tasks, a training dataset is a collection of triplets of query $\mathbf{q}$, passage $\mathbf{p}$, and answer $\mathbf{a}$. Say $\langle q_n, p_n, a_n\rangle$ is the $n$-th training instance. 

The first step is to extract memory $M$ from $p_n$ by mapping each symbolic in the passage to a contextual representation given by the concatenation of forward and backward RNN hidden states, i.e., $m_k = [\overrightarrow{p_{n}}^k, \overleftarrow{p_n}^{|p_n|-k + 1}]$, and extract initial state $s_1$ from $q_n$ by assigning $s_1 = [ \overrightarrow{q_n}^{|q_n|}, \overleftarrow{q_n}^1]$.    
Given $M$ and $s_1$ for the $n$-th training instance, a ReasoNet executes $|\mathbb{A}^\dagger|$ episodes, where all possible episodes $\mathbb{A}^\dagger$ can be enumerated by setting a maximum step. Each episode generates actions and a reward from the last step: $\langle (t_{1:T}, a_{T})$, $r_{T}\rangle_{(t_{1:T}, a_T) \in \mathbb{A}^\dagger}$.
Therefore, the gradient of $J$ can be rewritten as:
\[ \nabla_{\theta} J(\theta) = \sum_{(t_{1:T}, a_T) \in \mathbb{A}^\dagger} \pi(t_{1:T}, a_T; \theta) \left[ \nabla_{\theta} \text{log} \pi(t_{1:T}, a_{T}; \theta) ( r_{T} - b) \right]   \]
where the baseline $ b =  \sum_{(t_{1:T}, a_T) \in \mathbb{A}^\dagger} \pi(t_{1:T}, a_T; \theta) r_{T} $ is the average reward on the $|\mathbb{A}^\dagger|$ episodes for the $n$-th training instance.
It allows different baselines for different training instances. This can be beneficial since the complexity of training instances varies significantly. 
In experiments, we empirically find using $( \frac{r_{T}} { b} -1 )$ in replace of  $( r_{T} - b )$ can lead to a faster convergence. Therefore, we adopt this approach to train ReasoNets in the experiments.

\section{Experiments}
\label{sec:exp}
In this section, we evaluate the performance of ReasoNets in machine comprehension datasets, including unstructured CNN and Daily Mail datasets, the Stanford SQuAD dataset, and a structured Graph Reachability dataset.

\subsection{CNN and Daily Mail Datasets}

We examine the performance of ReasoNets on CNN and Daily Mail datasets.\footnote{The CNN and Daily Mail datasets are available at https://github.com/deepmind/rc-data}
The detailed settings of the ReasoNet model are as follows.

\textbf{Vocab Size}: For training our ReasoNet, we keep the most frequent $|{V}| = 101k$ words (not including 584 entities and 1 placeholder marker) in the CNN dataset, and $|{V}| = 151k$ words (not including 530 entities and 1 placeholder marker) in the Daily Mail dataset.

\textbf{Embedding Layer}: We choose $300$-dimensional word embeddings, and use the $300$-dimensional pretrained Glove word embeddings \citep{GloveEMNLP2014} for initialization. We also apply dropout with probability $0.2$ to the embedding layer.

\textbf{Bi-GRU Encoder}: We apply bidirectional GRU for encoding query and passage into vector representations. We set the number of hidden units to be $256$ and $384$ for the CNN and Daily Mail datasets, respectively. The recurrent weights of GRUs are initialized with random orthogonal matrices. The other weights in GRU cell are initialized from a uniform distribution between $-0.01$ and $0.01$. We use a shared GRU model for both query and passage.

\textbf{Memory and Attention}: The memory of the ReasoNet on CNN and Daily Mail dataset is composed of query memory and passage memory. $M = (M^{query}, M^{doc})$, where $M^{query}$ and $M^{doc}$ are extracted from query bidirectional-GRU encoder and passage bidirectional-GRU encoder respectively. We  choose projected cosine similarity function as the attention module. The attention score $a^{doc}_{t,i}$ on memory $m^{doc}_i$ given the state $s_t$ is computed as follows:
	  $ a^{doc}_{t,i} = \text{softmax}_{i=1,...,|M^{doc}|} \gamma \cos({W^{doc}_1} m^{doc}_i, {W^{doc}_2} s_t)$, 
where $\gamma$ is set to 10. ${W^{doc}_1}$ and ${W^{doc}_2}$ are weight vectors associated with $m^{doc}_i$ and $s_t$, respectively, and are joint trained in the ReasoNet. Thus, the attention vector on passage is given by $x^{doc}_t = \sum_{i}^{|M^{doc}|} a^{doc}_{t,i} m^{doc}_i$. Similarly, the attention vector on query is  $x^{query}_t = \sum_{i}^{|M^{query}|} a^{query}_{t,i} m^{query}_i$.
The final attention vector is the concatenation of the query attention vector and the passage attention vector $x_t = (x^{query}_t, x^{doc}_t)$. The attention module is parameterized by $\theta_x = ({W^{query}_1}, {W^{query}_2}, {W^{doc}_1}, {W^{doc}_2})$;

\begin{table}
	\centering
	{\small 	
		\caption{{The performance of Reasoning Network on CNN and Daily Mail dataset.}}
		\label{tab:cnn_dm_table}
		\begin{tabular}{lcccc}
			\toprule
			& \multicolumn{2}{c}{CNN}  &  \multicolumn{2}{c}{Daily Mail} \\
			\cmidrule{1-5}
			& valid & test & valid & test \\
			{Deep LSTM Reader \citep{HermannNIPS2015}} & 55.0 & 57.0 & 63.3 & 62.2 \\    
			{Attentive Reader \citep{HermannNIPS2015}} & 61.6 & 63.0 & 70.5 & 69.0 \\
			{MemNets \citep{HillICLR2016}} 		  & 63.4 & 66.8 & -    & - 	  \\
			{AS Reader \citep{KadlecAttentionSum2016}} 	      & 68.6 & 69.5 & 75.0 & 73.9 \\
			{Stanford AR \citep{ChenACL2016}}      & 72.2 & 72.4 & 76.9 & 75.8 \\ 
			{DER Network \citep{Kobayashi2016a}} & 71.3 & 72.9 & - & - \\
			{Iterative Attention Reader \citep{Sordoni2016IterativeAtt}} & 72.6 & 73.3 & - & -  \\
			{EpiReader \citep{Trischler2016EpiReader}} & \textbf{73.4} & 74.0 & - & - \\
			{GA Reader \citep{Dhingra2016Gated}} & 73.0 & 73.8 &  76.7 & 75.7  \\
			{AoA Reader \citep{Cui2016AoA}} & 73.1 & 74.4 & - & - \\
			{ReasoNet} &  72.9 & \textbf{74.7} & \textbf{77.6} & \textbf{76.6}  \\    
			\bottomrule
		\end{tabular}
	}
\end{table}

\begin{figure}[th!]
	\centering
	\includegraphics[width=0.4\textwidth]{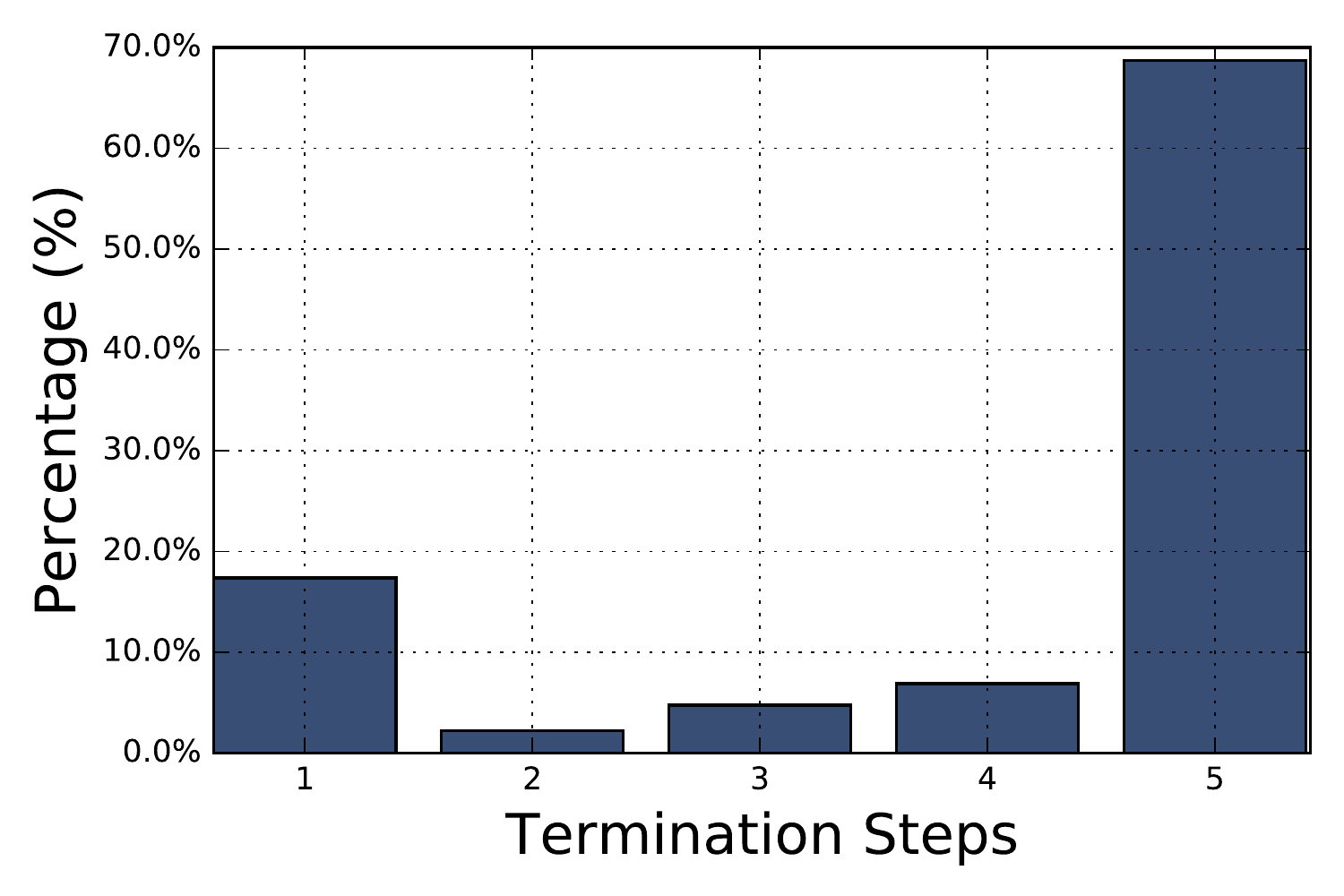}
	\caption{{The termination step distribution of a ReasoNet ($T_{max}=5$) in the CNN dataset.}}
	\label{fig:cnntsdistr}
\end{figure}

\textbf{Internal State Controller}: We choose GRU model as the internal state controller. The number of hidden units in the GRU state controller is $256$ for CNN and $384$ for Daily Mail. The initial state of the GRU controller is set to be the last-word of the query representation by a bidirectional-GRU encoder.

\begin{figure*}[th!]  
	\centering
	\includegraphics[width=0.6\textwidth]{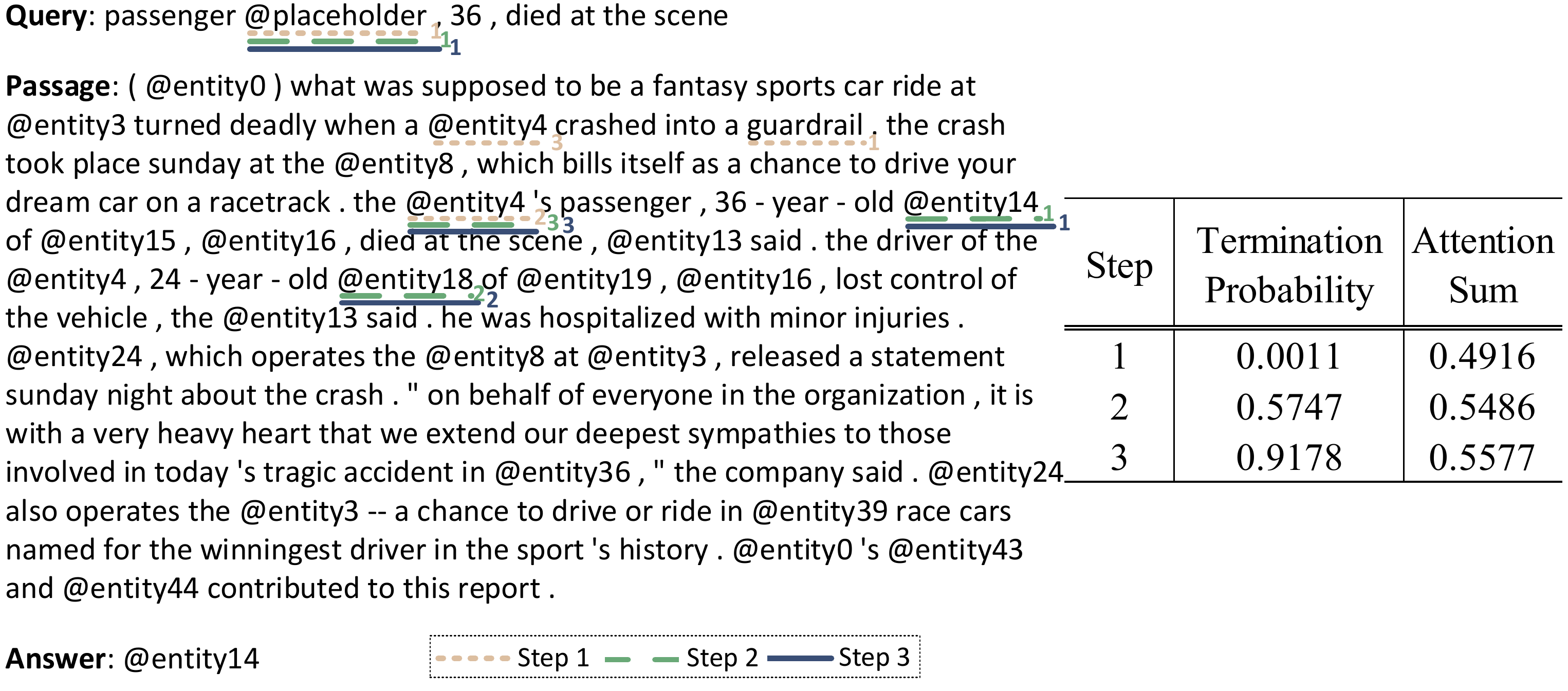}
	\caption{{Results of a test example 69e1f777e41bf67d5a22b7c69ae76f0ae873cf43.story from the CNN dataset. The numbers next to the underline bars indicate the rank of the attention scores. The corresponding termination probability and the sum of attention scores for the answer entity are shown in the table on the right.}}
	\label{fig:cnn_visualization}
\end{figure*}

\textbf{Termination Module}: We adopt a logistical regression to model the termination variable at each time step: 
\[f_{tg}(s_t; \theta_{tg}) = \text{sigmoid}({W_{tg}} s_t + {b_{tg}}); \theta_{tg} = ({W_{tg}}, {b_{tg}})\]
where $W_{tg}$ and $b_{tg}$ are the weight matrix and bias vector, respectively.

\textbf{Answer Module}: We apply a linear projection from GRU outputs and make predictions on the entity candidates. Following the settings in AS Reader \citep{KadlecAttentionSum2016}, we sum up scores from the same candidate and make a prediction. Thus, AS Reader can be viewed as a special case of ReasoNets with $T_{\text{max}} = 1$.\footnote{When ReasoNet is set with $T_{\text{max}} = 1$ in CNN and Daily Mail, it directly applies $s_0$ to make predictions on the entity candidates, without performing attention on the memory module. The prediction module in ReasoNets is the same as in AS Reader. It sums up the scores from the same entity candidates, where the scores are calculated by the inner product between $s_t$ and $m^{doc}_{e}$, where $m^{doc}_{e}$ is an embedding vector of one entity candidate in the passage.}

\textbf{Other Details}: The maximum reasoning step, $T_{\text{max}}$ is set to $5$ in experiments on both CNN and Daily Mail datasets. We use ADAM optimizer \citep{KingmaICLR2015Adam} for parameter optimization with an initial learning rate of $0.0005$, $\beta_1 = 0.9$ and $\beta_2 = 0.999$; The absolute value of gradient on each parameter is clipped within $0.001$. The batch size is 64 for both CNN and Daily Mail datasets. For each batch of the CNN and Daily Mail datasets, we randomly reshuffle the assignment of named entities \citep{HermannNIPS2015}.
This forces the model to treat the named entities as semantically meaningless labels. In the prediction of test cases, we randomly reshuffle named entities up to 4 times, and report the averaged answer. Models are trained on GTX TitanX 12GB. It takes 7 hours per epoch to train on the Daily Mail dataset and 3 hours per epoch to train on the CNN dataset. The models are usually converged within 6 epochs on both CNN and Daily Mail datasets.

\textbf{Results}: Table \ref{tab:cnn_dm_table} shows the performance of all the existing single model baselines and our proposed ReasoNet. Among all the baselines, AS Reader could be viewed as a special case of ReasoNet with $T_{\text{max}} = 1$. Comparing with the AS Reader, ReasoNet shows the significant improvement by capturing multi-turn reasoning in the paragraph. Iterative Attention Reader, EpiReader and GA Reader are the three multi-turn reasoning models with fixed reasoning steps. ReasoNet also outperforms all of them by integrating termination gate in the model which allows different reasoning steps for different test cases. AoA Reader is another single-turn reasoning model, it captures the word alignment signals between query and passage, and shows a big improvement over AS Reader. ReasoNet obtains comparable results with AoA Reader on CNN test set. We expect that ReasoNet could be improved further by incorporating the word alignment information in the memory module as suggested in AoA Reader. 

We show the distribution of termination step distribution of ReasoNets in the CNN dataset in Figure \ref{fig:cnntsdistr}. The distributions spread out across different steps. Around 70\% of the instances terminate in the last step.
Figure \ref{fig:cnn_visualization} gives a test example on CNN dataset, which illustrates the inference process of the ReasoNet. The model initially focuses on wrong entities with low termination probability. In the second and third steps, the model focuses on the right clue with higher termination probability. Interestingly, we also find its query attention focuses on the placeholder token throughout all the steps. 

\begin{table}
	\centering
	{\small 	
		\caption{{Results on the SQuAD test leaderboard.}}
		\label{tab:squad_results}
		\begin{tabular}{lcccc}
			\toprule
			& \multicolumn{2}{c}{Single Model}  &  \multicolumn{2}{c}{Ensemble Model} \\
			\cmidrule{1-5}
			& EM & F1 & EM & F1 \\
			{Logistic Regression Baseline \citep{Squad2016}} & 40.4 & 51.0 & - & - \\    
			{Dynamic Chunk Reader \citep{DBLP:journals/corr/YuZHYXZ16}} & 62.5 & 71.0 & - & - \\
			{Fine-Grained Gating \citep{DBLP:journals/corr/YangDYHCS16}} 		  & 62.5 & 73.3 & -    & - 	  \\
			{Match-LSTM \citep{DBLP:journals/corr/WangJ16a}} 	      & 64.7 & 73.7 & 67.9 & 77.0 \\
			{RaSoR \citep{DBLP:journals/corr/LeeKP016}} 			 &  - &  - & 67.4 & 75.5  \\
			{Multi-Perspective Matching \citep{IBM_MC_2016}}      & 68.9 & 77.8 & 73.7 & 81.3 \\ 
			{Dynamic Coattention Networks \citep{DBLP:journals/corr/XiongZS16}} & 66.2 & 75.9 & 71.6 & 80.4 \\
			{BiDAF \citep{Dhingra2016Gated}} & 68.0 & 77.3 & 73.3 & 81.1  \\
			{ReasoNet} &  \textbf{69.1} & \textbf{78.9} & \textbf{73.4} & \textbf{81.8} \\	
			\cmidrule{1-5}			
			{Iterative Co-attention Network$^{\alpha}$} & 67.5 & 76.8 & - & -  \\
			{FastQA \citep{DBLP:journals/corr/WeissenbornWS17}} &  68.4 & 77.1 & 70.8 & 78.9 \\
			
			{jNet \citep{DBLP:journals/corr/ZhangZCDWJ17}} & 68.7 & 77.4 & - & - \\
			{Document Reader \cite{DBLP:journals/corr/ChenFWB17}} &  69.9 & 78.9 & - & -  \\
			{R-Net \citep{rnet}} &  \textbf{71.3} & \textbf{79.7} & \textbf{75.9} & \textbf{82.9}   \\
			\bottomrule
		\end{tabular}
	}
\end{table}

\begin{table*}[th]
	\centering
	{\small
		\caption{{Reachability statistics of the Graph Reachability dataset.}}
		\label{tab:statgraph}
		\vspace{-1mm}
		\begin{tabular}{l|cccccccc}
			\toprule
			& \multicolumn{4}{c}{Small Graph}  & \multicolumn{4}{c}{Large Graph}                \\
			\midrule
			Reachable Step & No Reach  &  1--3 & 4--6 & 7--9 & No Reach  &  1--3 & 4--6 & 7--13	\\\hline
			Train  ($\%$) 	  & $44.16$ &  $42.06$ & $13.51$  & $0.27$ & $49.02$ & $25.57$ & $21.92$  & $3.49$ \\
			Test   ($\%$)     & $45.00$ &  $41.35$ & $13.44$ & $0.21$ & $49.27$ &  $25.46$ & $21.74$  & $3.53$  \\
			
			\bottomrule
		\end{tabular}
	}
\end{table*}
\begin{table*}[ht]
	{\small
		\caption{{Small and large random graph in the Graph Reachability dataset. Note that ``$A \rightarrow B$'' represents an edge connected from $A$ to $B$ and the \# symbol is used as a delimiter between different edges.}}
		\vspace{-1mm}
		\label{tab:sample-graph}
		\centering
		\begin{tabular}{lll}
			\toprule
			& \multicolumn{1}{c}{Small Graph}  &   \multicolumn{1}{c}{Large Graph}               \\
			\cmidrule{1-3}
			Graph Description  
			& 0 $\rightarrow$ 0 \# 0 $\rightarrow$ 2 \# 1 $\rightarrow$ 2 \# 2 $\rightarrow$ 1 \# &  0 $\rightarrow$ 17 \#  1 $\rightarrow$ 3 \# 1 $\rightarrow$ 14 \# 1 $\rightarrow$ 6 \# \\ 
			& 3 $\rightarrow$ 2 \# 3 $\rightarrow$ 3 \# 3 $\rightarrow$ 6 \# 3 $\rightarrow$ 7 \# & 2 $\rightarrow$ 11 \#  2 $\rightarrow$ 13 \# 2 $\rightarrow$ 15 \# 3 $\rightarrow$ 7\# \\ 	        
			& 4 $\rightarrow$ 0 \# 4 $\rightarrow$ 1 \# 4 $\rightarrow$ 4 \# 5 $\rightarrow$ 7 \#  & 5 $\rightarrow$ 0  \#  5 $\rightarrow$ 7 \# 6 $\rightarrow$ 10 \# 6 $\rightarrow$ 5\# \\
			& 6 $\rightarrow$ 0 \# 6 $\rightarrow$ 1 \# 7 $\rightarrow$ 0 \#  &  7 $\rightarrow$ 15 \# 7 $\rightarrow$ 7 \# 8 $\rightarrow$ 11 \# 8 $\rightarrow$ 7 \# \\ 
			& & 10 $\rightarrow$ 9  \# 10 $\rightarrow$ 6 \# 10 $\rightarrow$ 7 \# 12 $\rightarrow$ 1 \#  \\ 
			& & 12 $\rightarrow$ 12 \# 12 $\rightarrow$ 6 \# 13 $\rightarrow$ 11 \# 14 $\rightarrow$ 17 \# \\ 
			& & 14 $\rightarrow$ 14  \# 15 $\rightarrow$ 10 \#  16 $\rightarrow$ 2 \# 17 $\rightarrow$ 4 \# \\ 
			& & 17 $\rightarrow$ 7 \#  \\
			\midrule
			Query &  7 $\rightarrow$ 4  & 10 $\rightarrow$ 17 \\\hline
			Answer     & No  & Yes \\
			\bottomrule
		\end{tabular}
	}
\end{table*}
\subsection{SQuAD Dataset}
In this section, we evaluate ReasoNet model on the task of question answering using the SQuAD dataset \citep{Squad2016}.\footnote{SQuAD Competition Website is https://rajpurkar.github.io/SQuAD-explorer/} SQuAD is a machine comprehension dataset on 536 Wikipedia articles, with
more than 100,000 questions. Two metrics are used to evaluate models: Exact Match (EM) and a softer metric, F1 score, which measures the weighted average of the precision and recall rate at the character level. The dataset consists of 90k/10k training/dev question-context-answer tuples with a large hidden test set. The model architecture used for this task is as follows:

\textbf{Vocab Size}: We use the python NLTK tokenizer\footnote{NLTK package could be downloaded from http://www.nltk.org/} to preprocess passages and questions, and obtain about 100K words in the vocabulary.   

\textbf{Embedding Layer}: We use the $100$-dimensional pretrained Glove vectors \citep{GloveEMNLP2014} as word embeddings. These Glove vectors are fixed during the model training. To alleviate the out-of-vocabulary issue, we adopt one layer $100$-dimensional convolutional neural network on character-level with a width size of $5$ and each character encoded as an $8$-dimensional vector following the work \citep{DBLP:journals/corr/SeoKFH16}. The $100$-dimensional Glove word vector and the $100$-dimensional character-level vector are concatenated to obtain a $200$-dimensional vector for each word.

\textbf{Bi-GRU Encoder}: We apply bidirectional GRU for encoding query and passage into vector representations. The number of hidden units is set to 128.

\textbf{Memory}: We use bidirectional-GRU encoders to extract the query representation $M^{query}$ and the passage representation $M^{doc}$, given a query and a passage. 
We compute the similarity matrix between each word in the query and each word in the passage. The similarity matrix is denoted as $S \in \mathbb{R}^{\mathcal{T} \times J}$, where $\mathcal{T}$ and $J$ are the number of words in the passage and query, respectively, and $S_{tj} = w_{S}^\intercal  [ M^{doc}_{:t}; M^{query}_{:j}; M^{doc}_{:t} \circ M^{query}_{:j} ] \in \mathbb{R} $, where $w_S$ is a trainable weight vector, $\circ$ denotes the elementwise multiplication,
and $[;]$ is the vector concatenation across row.
We then compute the context-to-query attention and query-to-context attention from the similarity matrix $S$ by following recent co-attention work \citep{DBLP:journals/corr/SeoKFH16} to obtain the query-aware passage representation $G$. We feed $G$ to a $128$-dimensional bidirectional GRU to obtain the memory $M = \text{bidirectional-GRU}(G)$, where $M \in \mathbb{R}^{256 \times \mathcal{T}}$. 

\textbf{Internal State Controller}: We use a GRU model with $256$-dimensional hidden units as the internal state controller. The initial state of the GRU controller is the last-word representation of the query bidirectional-GRU encoder.

\textbf{Termination Module}: We use the same termination module as in the CNN and Daily Mail experiments.

\textbf{Answer Module}: SQuAD task requires the model to find a span in the passage to answer the query. Thus the answer module requires to predict the start and end indices of the answer span in the passage. The probability distribution of selecting the start index over the passage at state $s_t$ is computed by :
\[
p^{1}_{t} = \text{softmax}( w^{\intercal}_{p^{1}} [M; M \circ S_t] )
\]
where $S_t$ is given via tiling $s_t$ by $T$ times across the column and  $w_{p^{1}}$ is a trainable weight vector. The probability distribution of selecting the end index over passage is computed in a similar manner:
\[
p^{2}_{t} = \text{softmax}( w^{\intercal}_{p^{2}} [M; M \circ S_t] )
\]

\textbf{Other Details}: The maximum reasoning step $T_{\text{max}}$ is set to $10$ in SQuAD experiments. We use AdaDelta optimizer \citep{DBLP:journals/corr/abs-1212-5701} for parameter optimization with an initial learning rate of $0.5$ and a batch size of 32. Models are trained on GTX TitanX 12GB. It takes about 40 minutes per epoch for training, with 18 epochs in total.

\textbf{Results} :
In the Table \ref{tab:squad_results}, we report the performance of all models in the SQuAD leaderboard.\footnote{Results shown here reflect the SQuAD leaderboard (stanford-qa.com) as of 17 Feb 2017, 9pm PST. We include the reference in the camera-ready version. $\alpha:$ Fudan University.} In the upper part of the Table \ref{tab:squad_results}, we compare ReasoNet with all published baselines at the time of submission. Specifically, BiDAF model could be viewed as a special case of ReasoNet with $T_{max} = 1$. It is worth noting that this SQuAD leaderboard is highly active and competitive. The test set is hidden to all models and all the results on the leaderboard are produced and reported by the organizer; thus all the results here are reproducible. In Table \ref{tab:squad_results}, we demonstrate that ReasoNet outperforms all existing published approaches. While we compare ReasoNet with BiDAF, ReasoNet exceeds BiDAF both in single model and ensemble model cases.  This demonstrates the importance of the dynamic multi-turn reasoning over a passage. In the bottom part of Table \ref{tab:squad_results}, we compare ReasoNet with all unpublished methods at the time of this submission, ReasoNet holds the second position in all the competing approaches in the SQuAD leaderboard.

\begin{table*}
	\centering
	{\small
		\caption{The performance of Reasoning Network on the Graph Reachability dataset.}
		\label{tab:reach_table}
		\centering
		\begin{tabular}{lllllll}
			\toprule
			& \multicolumn{3}{c}{Small Graph}  &   \multicolumn{3}{c}{Large Graph}               \\
			\cmidrule{1-7}
			& ROC-AUC & PR-AUC & Accuracy & ROC-AUC & PR-AUC & Accuracy \\
			{Deep LSTM Reader} & 0.9619 & 0.9565 & 0.9092 & 0.7988 & 0.7887 & 0.7155 \\
			{ReasoNet-$T_\text{max}=2$} & 0.9638 & 0.9677 & 0.8961 & 0.8477 & 0.8388 & 0.7607 \\
			{ReasoNet-Last} & \textbf{1} & \textbf{1} & \textbf{1} & 0.8836 & 0.8742 & 0.7895 \\
			
			{ReasoNet} & \textbf{1} & \textbf{1} & \textbf{1} & \textbf{0.9988} & \textbf{0.9989} & \textbf{0.9821} \\    
			\bottomrule
		\end{tabular}
	}
\end{table*}

\begin{figure*}[h!]
	\centering
	\includegraphics[width=0.5\textwidth]{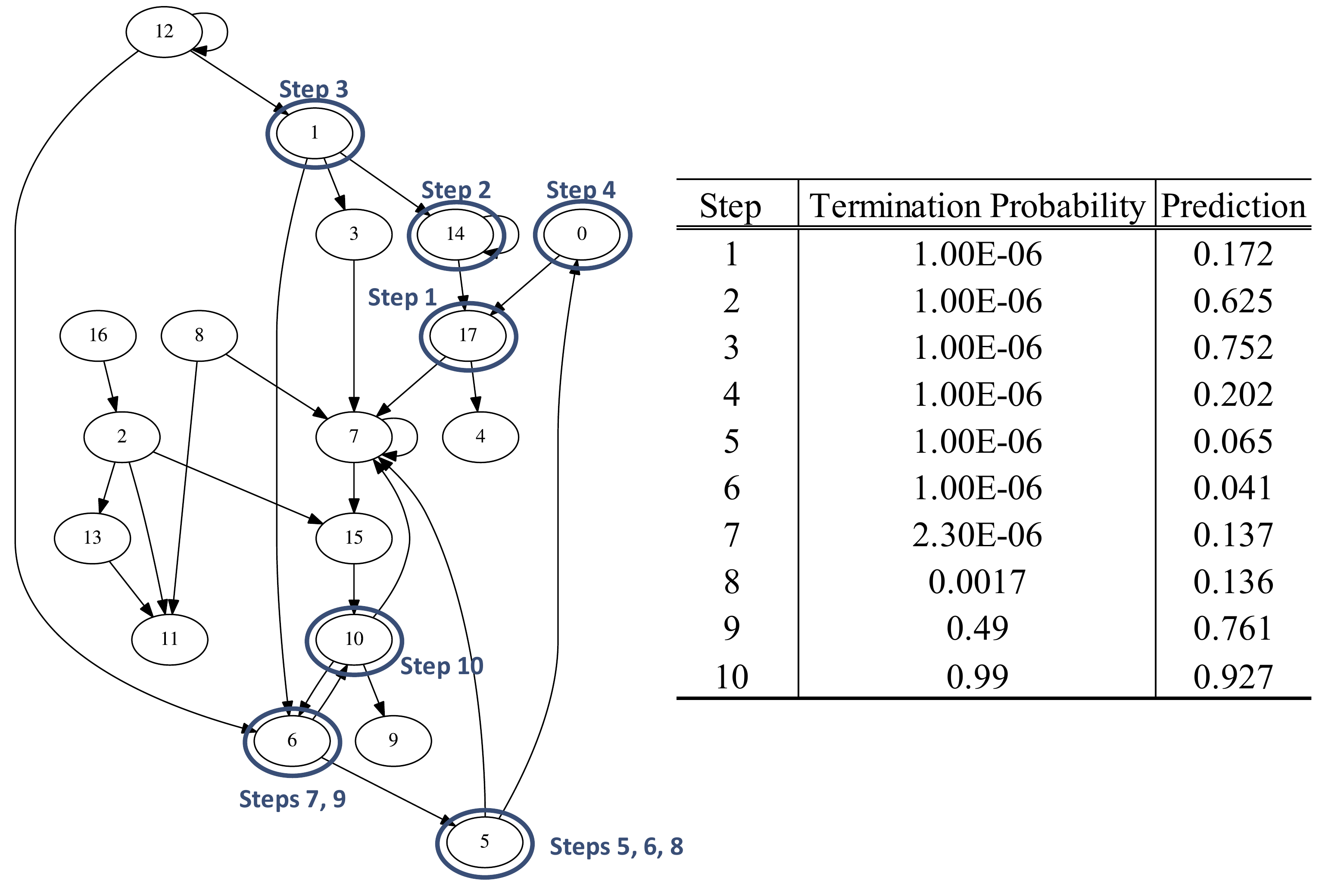}
	\caption{{An example of graph reachability result, given a query ``10 $\rightarrow$ 17'' (Answer: Yes). The red circles highlight the nodes/edges which have the highest attention in each step. The corresponding termination probability and prediction results are shown in the table. The model terminates at step $10$.}}
	\label{fig:graph_visalization}
\end{figure*}

\subsection{Graph Reachability Task}
Recent analysis and results \citep{ChenACL2016} on the cloze-style machine comprehension tasks have suggested some simple models without multi-turn reasoning can achieve reasonable performance. Based on these results, we construct a synthetic structured Graph Reachability dataset\footnote{The dataset is available at https://github.com/MSRDL/graph\_reachability\_dataset} to evaluate longer range machine inference and reasoning capability, since we anticipate ReasoNets to have the capability to handle long range relationships.
 
We generate two synthetic datasets: a small graph dataset and a large graph dataset. In the small graph dataset, it contains $500 K$ small graphs, where each graph contains $9$ nodes and $16$ direct edges to randomly connect pairs of nodes. The large graph dataset contains $500 K$ graphs, where each graph contains $18$ nodes and $32$ random direct edges. Duplicated edges are removed. Table \ref{tab:statgraph} shows the graph reachability statistics on the two datasets. 

\begin{figure*}[t!]
	\centering
	\includegraphics[width=0.5\textwidth]{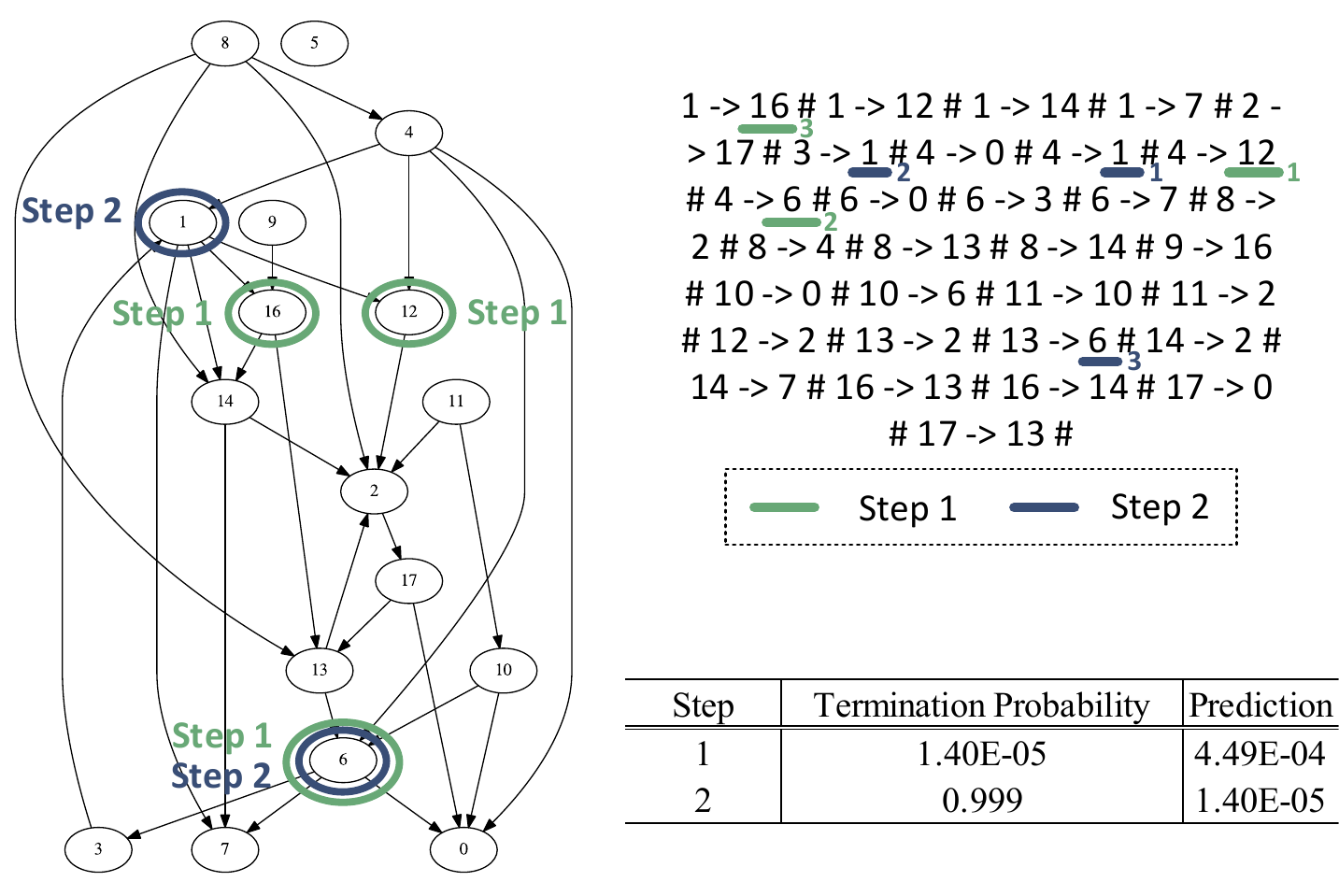}
	\caption{{An example of graph reachability result, given a query ``4 $\rightarrow$ 9'' (Answer: No).  The numbers next to the underline bars indicate the rank of the attention scores. The corresponding termination probability and prediction results are shown in the table.}}
	\label{fig:graph_visalization_fail}
\end{figure*}

In Table \ref{tab:sample-graph}, we show examples of a small graph and a large graph in the synthetic dataset. Both graph and query are represented by a sequence of symbols. The details settings of the ReasoNet are listed as follows in the reachability tasks.

\textbf{Embedding Layer} We use a $100$-dimensional embedding vector for each symbol in the query and graph description.

\textbf{Bi-LSTM Encoder}: We apply a bidirectional-LSTM layer with $128$ and $256$ cells on query embeddings in the small and large graph datasets, respectively. The last states of bidirectional-LSTM on query are concatenated to be the initial internal state $s_1 = [ \overrightarrow{q}^{|q|}, \overleftarrow{q}^1]$ in the ReasoNet.

\textbf{Memory}: We apply another bidirectional-LSTM layer with $128$ and $256$ cells on graph description embeddings in the small and large graph datasets, respectively. It maps each symbol $g^i$ to a contextual representation given by the concatenation of forward and backward LSTM hidden states $m_i = [\overrightarrow{g}^{i}, \overleftarrow{g}^{|g|-i + 1}]$.

\textbf{Internal State Controller}: We use a GRU model with $128$-dimensional and $256$-dimensional hidden units as the internal state controller for the small and large graph datasets, respectively. The initial state of the GRU controller is $s_1$.

\textbf{Answer Module}: The final answer is either ``Yes'' or ``No'' and hence logistical regression is used as the answer module:  $a_t = \sigma({W_a} s_t + {b_a})$; $\theta_{a} = ({W_{a}}, {b_{a}})$.

\textbf{Termination Module}: We use the same termination module as in the CNN and Daily Mail experiments.

\textbf{Other Details}: The maximum reasoning step $T_{\text{max}}$ is set to $15$ and $25$ for the small graph and large graph dataset, respectively. We use AdaDelta optimizer \citep{DBLP:journals/corr/abs-1212-5701} for parameter optimization with an initial learning rate of $0.5$ and a batch size of 32.

We denote ``\texttt{ReasoNet}'' as the standard ReasoNet with termination gate, as described in Section \ref{sec:train_detail}. To study the effectiveness of the termination gate in ReasoNets, we remove the termination gate and use the prediction from the last state, $\hat{a} = a_{T_{\text{max}}}$ ($T_{\text{max}}$ is the maximum reasoning step), denoted as ``\texttt{ReasoNet-Last}''. To study the effectiveness of multi-turn reasoning, we choose ``\texttt{ReasoNet-$T_\text{max}=2$}'', which only has single-turn reasoning.
We compare ReasoNets with a two layer deep LSTM model \citep{HermannNIPS2015} with $128$ hidden units, denoted as ``\texttt{Deep LSTM Reader}'', as a baseline.
Table \ref{tab:reach_table} shows the performance of these models on the graph reachability dataset.
\texttt{Deep LSTM Reader} achieves $90.92\%$ and $71.55\%$ accuracy in the small and large graph dataset, respectively, which indicates the graph reachibility task is not trivial.
The results of \texttt{ReasoNet-$T_\text{max}=2$} are comparable with the results of \texttt{Deep LSTM Reader}, since both \texttt{Deep LSTM Reader} and \texttt{ReasoNet-$T_\text{max}=2$} perform single-turn reasoning.
The \texttt{ReasoNet-Last} model achieves $100\%$ accuracy on the small graph dataset, while the \texttt{ReasoNet-Last} model achieves only $78.95\%$ accuracy on the large graph dataset, as the task becomes more challenging.
Meanwhile, the \texttt{ReasoNet} model converges faster than the \texttt{ReasoNet-Last} model. The \texttt{ReasoNet} model converges in 20 epochs in the small graph dataset, and 40 epochs in the large graph dataset, while the \texttt{ReasoNet-Last} model converges around 40 epochs in the small graph dataset, and 70 epochs in the large graph dataset. The results suggest that the termination gate variable in the ReasoNet is helpful when training with sophisticated examples, and makes models converge faster. Both the \texttt{ReasoNet} and \texttt{ReasoNet-Last} models perform better than the \texttt{ReasoNet-$T_\text{max}=2$} model, which demonstrates the importance of the multi-turn reasoning. 

To further understand the inference process in ReasoNets, Figures \ref{fig:graph_visalization} and \ref{fig:graph_visalization_fail} show test examples of the large graph dataset. In Figure \ref{fig:graph_visalization}, we can observe that the model does not make a firm prediction till step $9$. The highest attention word at each step shows the reasoning process of the model. Interestingly, the model starts from the end node ($17$), traverses backward till finding the starting node ($10$) in step $9$, and makes a firm termination prediction. On the other hand, in Figure \ref{fig:graph_visalization_fail}, the model learns to stop in step $2$. In step $1$, the model looks for neighbor nodes ($12$, $6$, $16$) to $4$ and $9$. Then, the model gives up in step $2$ and predict ``No". All of these demonstrate the dynamic termination characteristic and potential reasoning capability of ReasoNets.

\begin{figure}[t]
	\centering
	\begin{subfigure}{.44\textwidth}
		\centering
		\includegraphics[width=0.8\textwidth]{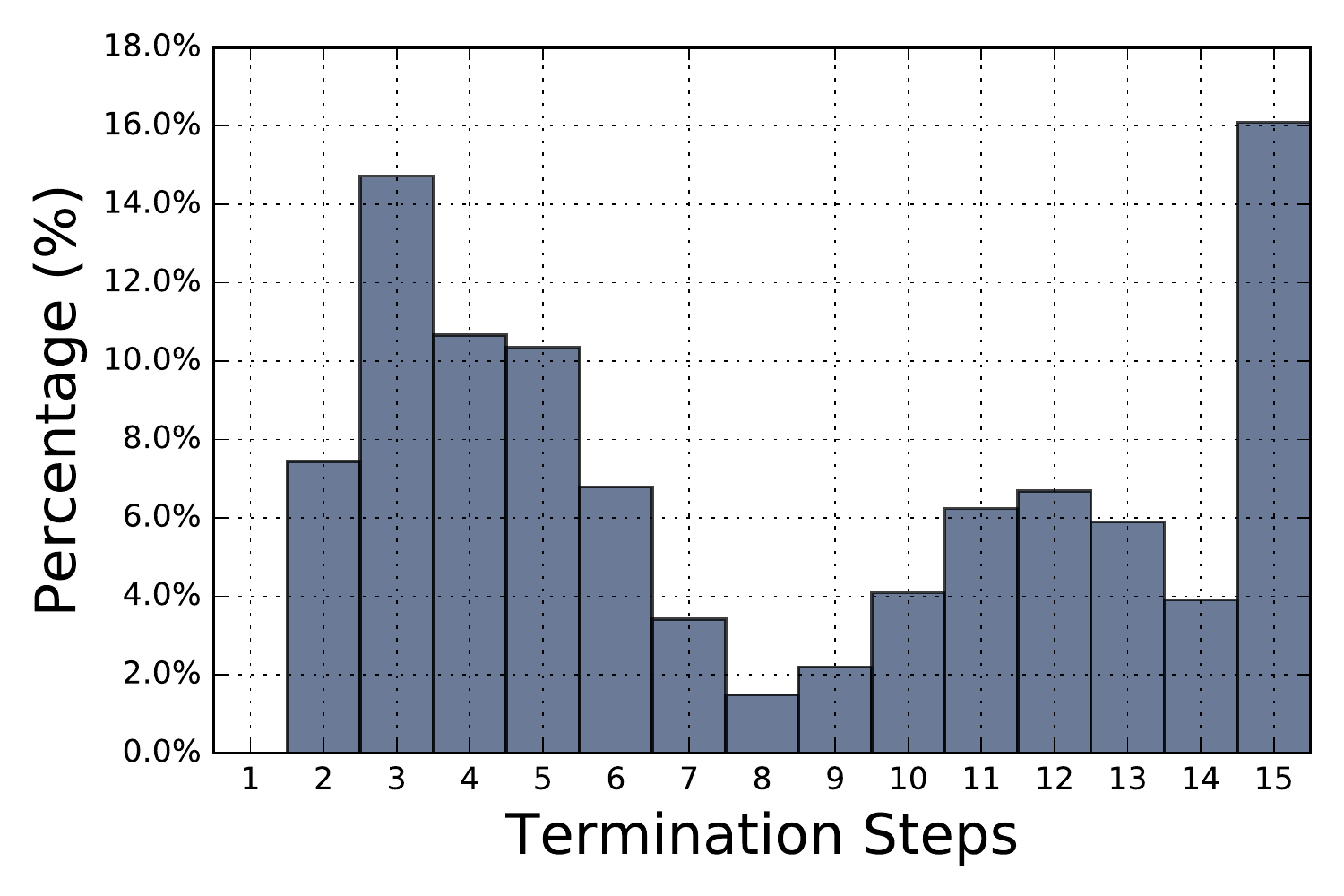}
		\caption{Small Graph}
	\end{subfigure}
	\begin{subfigure}{.44\textwidth}
		\centering
		\includegraphics[width=0.8\textwidth]{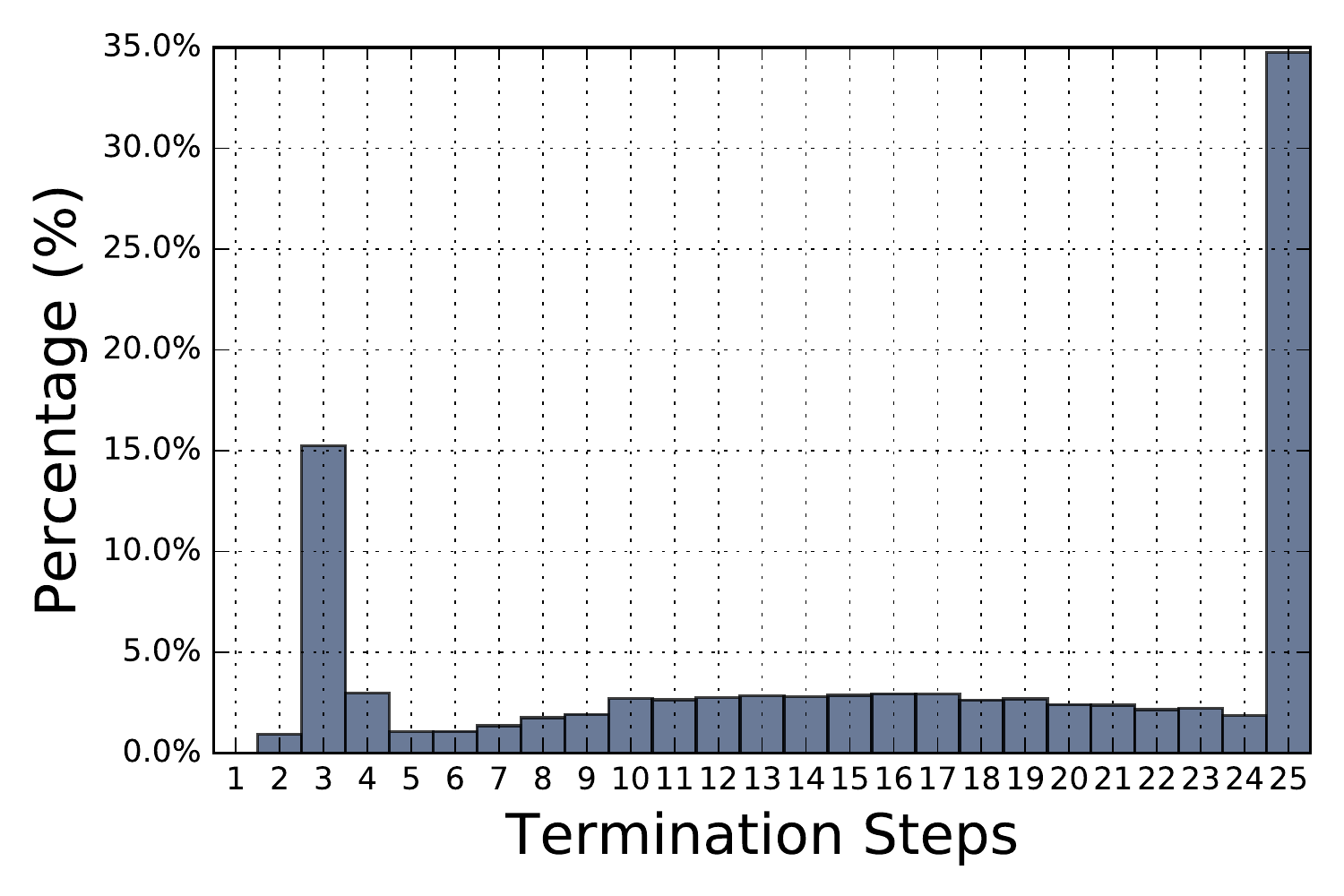} 
		\caption{Large Graph}
	\end{subfigure}
	\caption{{Termination step distribution of ReasoNets in the graph reachability dataset, where $T_{max}$ is set to $15$ and $25$ in the small graph and large graph dataset, respectively.}}
	\label{fig:graph_termination_histogram}
\end{figure}

\begin{figure}[t]
	\centering
	\begin{subfigure}{.41\textwidth}
		\centering
		\includegraphics[width=0.8\textwidth]{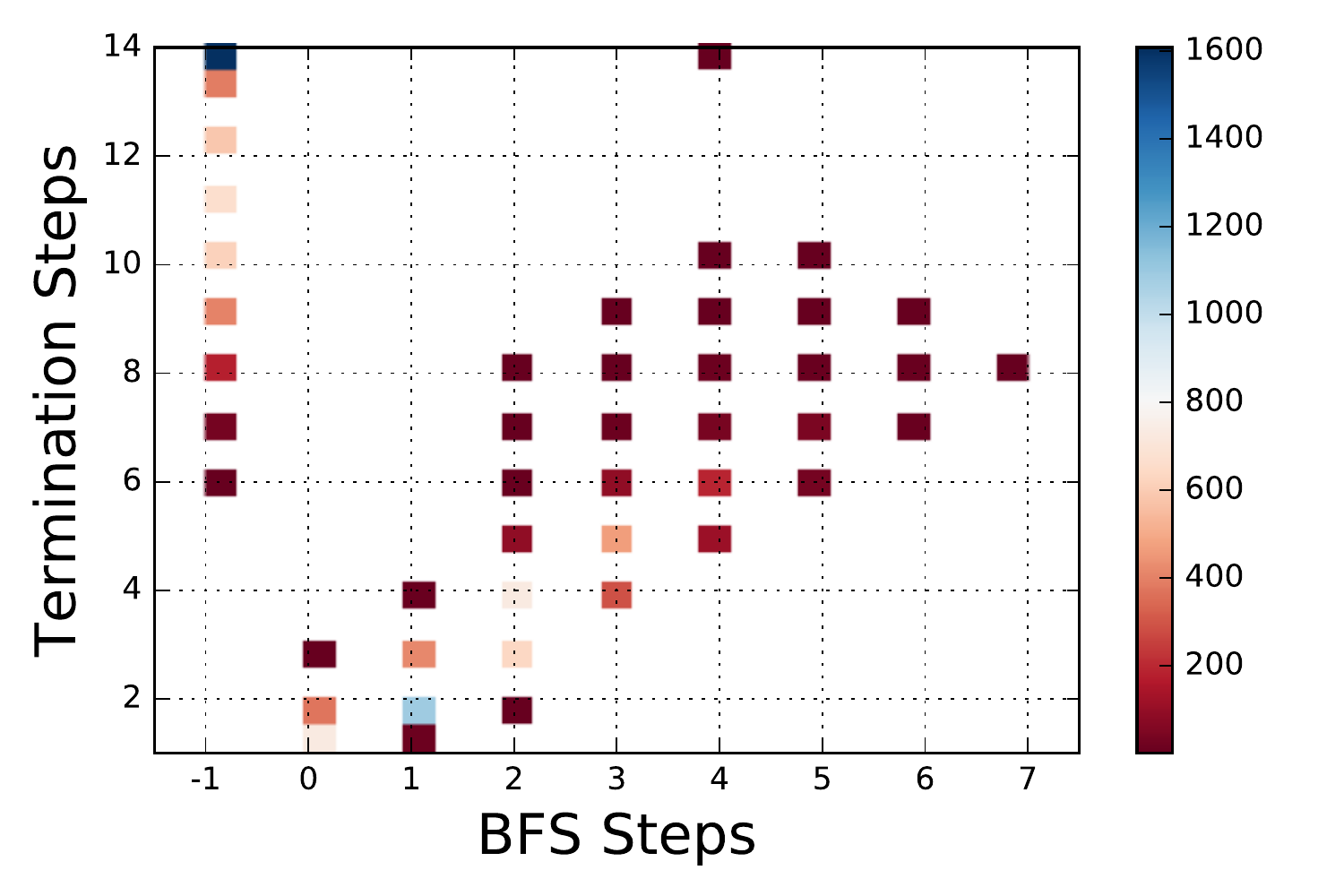}
			\vspace{-1mm}
		\caption{Small Graph}
	\end{subfigure}
	\begin{subfigure}{.41\textwidth}
		\centering
		\includegraphics[width=0.8\textwidth]{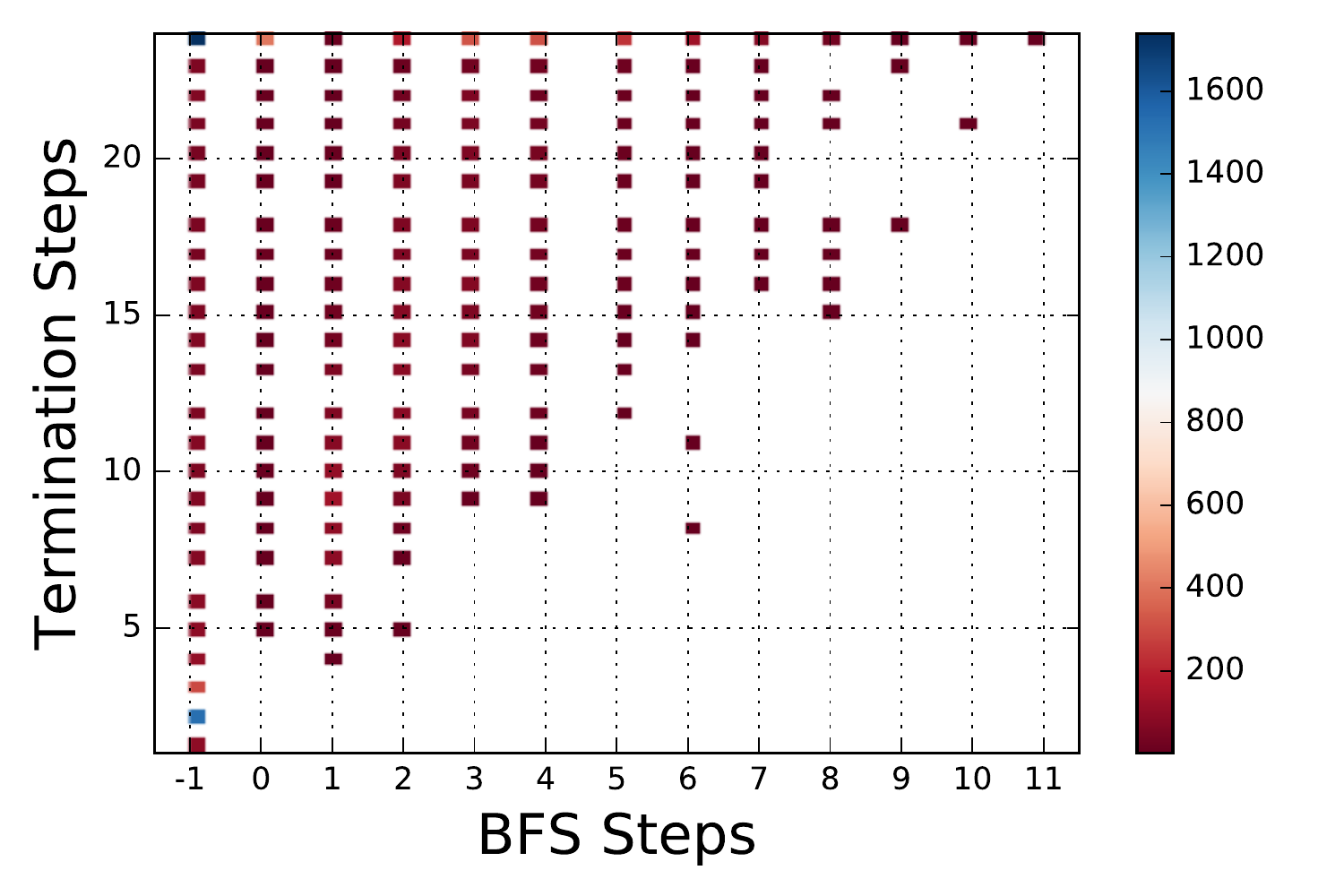}
	\vspace{-1mm}
		\caption{Large Graph}
	\end{subfigure}
	\vspace{-1mm}
	\caption{{The correlation between BFS steps and ReasoNet termination steps in the graph reachability dataset, where  $T_{max}$ is set to $15$ and $25$ in the small graph and large graph dataset, respectively, and BFS-Step$ =-1$ denotes unreachable cases. The value indicates the number of instances in each case.}}
	\label{fig:graph_termination_bfs_correlation}
\end{figure}

To better grasp when ReasoNets stop reasoning, we show the distribution of termination steps in ReasoNets on the test set. 
The termination step is chosen with the maximum termination probability $p(k) = t_k \prod_{i=1}^{k-1}{(1-t_i)}$, where $t_i$ is the termination probability at step $i$.
Figure \ref{fig:graph_termination_histogram} shows the termination step distribution of ReasoNets in the graph reachability dataset. 
The distributions spread out across different steps. Around 16\% and 35\% of the instances terminate in the last step for the small and large graph, respectively.
We study the correlation between the termination steps and the complexity of test instances in Figure \ref{fig:graph_termination_bfs_correlation}. 
Given the query, we use the Breadth-First Search (BFS) algorithm over the target graph to analyze the complexity of test instances.
For example, BFS-Step $ = 2$ indicates that there are two intermediate nodes in the shortest reachability path. Test instances with larger BFS-Steps are more challenging. We denote BFS-Step $ = -1$ as there is no reachable path for the given query. Figure \ref{fig:graph_termination_bfs_correlation} shows that test instances with larger BFS-Steps require more reasoning steps.

\section{Conclusion}
In this paper, we propose ReasoNets that dynamically decide whether to continue or to terminate the inference process in machine comprehension tasks.
With the use of the instance-dependent baseline method, our proposed model achieves superior results in machine comprehension datasets, including unstructured CNN and Daily Mail datasets, the Stanford SQuAD dataset, and a proposed structured Graph Reachability dataset.

\section{Acknowledgments}
We thank Ming-Wei Chang, Li Deng, Lihong Li, and Xiaodong Liu for their thoughtful feedback and discussions. 
%

{
\bibliography{reasonet_mrc}}
\bibliographystyle{ACM-Reference-Format}


\end{document}